\documentclass{article}

\PassOptionsToPackage{numbers, compress}{natbib}

\usepackage[preprint]{nips_2018}

\usepackage[utf8]{inputenc} 
\usepackage[T1]{fontenc}    
\usepackage{hyperref}       
\usepackage{url}            
\usepackage{booktabs}       
\usepackage{amsfonts}       
\usepackage{nicefrac}       
\usepackage{microtype}      
\usepackage{amsmath}
\usepackage{bm}
\usepackage{graphicx}
\usepackage{subfigure}
\usepackage{multicol}
\title{Tackling Early Sparse Gradients in Softmax Activation Using Leaky Squared Euclidean Distance}
\author{Wei Shen \qquad Rujie Liu \\
Fujitsu Research \& Development Center, Beijing, China.
\\
{\tt\small \{shenwei, rjliu\}@cn.fujitsu.com}
}

\begin{document}

\maketitle

\begin{abstract}
Softmax activation is commonly used to output the probability distribution over categories based on certain distance metric. In scenarios like one-shot learning, the distance metric is often chosen to be squared Euclidean distance between the query sample and the category prototype. This practice works well in most time. However, we find that choosing squared Euclidean distance may cause distance explosion leading gradients to be extremely sparse in the early stage of back propagation. We term this phenomena as the early sparse gradients problem. Though it doesn't deteriorate the convergence of the model, it may set up a barrier to further model improvement. To tackle this problem, we propose to use leaky squared Euclidean distance to impose a restriction on distances. In this way, we can avoid distance explosion and increase the magnitude of gradients. Extensive experiments are conducted on Omniglot and miniImageNet datasets. We show that using leaky squared Euclidean distance can improve one-shot classification accuracy on both datasets.
\end{abstract}

\section{Introduction}
Softmax activation is common in today's classification models~\cite{he2016deep}\cite{huang2017densely}\cite{krizhevsky2012imagenet}\cite{simonyan2014very}\cite{szegedy2015going}. It has several nice properties, such as probability interpretation of the prediction and inherent competition between different categories. 
State-of-the-art performances in tasks, such as face recognition~\cite{sun2014deep}, machine translation~\cite{vaswani2017attention}, large-scale image classification~\cite{he2016deep}, and few-shot classification ~\cite{snell2017prototypical} demonstrate the effectiveness of softmax activation.

Despite the success of softmax activation, researchers also start to pay attention to the activation itself~\cite{chen2017noisy}\cite{liu2017sphereface}\cite{liu2016large}\cite{wen2016discriminative}. While those works focus on improving softmax activation in parametric classification, we investigate its combination with nearest neighbor classifier which is typically used in non-parametric classification and is suitable for tasks like one-shot learning. 

When a nearest neighbor classifier is used for image classification, distances between the test image and category prototypes are calculated to measure their similarity. The classifier does not yield the probability of the prediction, but one can apply softmax activation over the negative distances to obtain the probability. The above calculation is often conducted in feature space and the choice of distance metric is usually squared Euclidean distance which is shown to outperform cosine distance in few-shot classification~\cite{snell2017prototypical}. 
However, the promising classification results based on squared Euclidean distance may hide the side effect of combining squared Euclidean distance with softmax activation. We observe that if the dimension of the feature vector is very large, it may cause squared Euclidean distances to explode which will push the negative log-softmax probability into either its saturation area or linear area (see area (III) and (I) in Figure~\ref{fig:pa_wrt_b}). When we use back propagation to update network weights, gradients of softmax loss with respect to the distances will be mostly 0 and -1, leading to the sparse gradients problem. Sparse gradients are undesired in the early training stage because it prevents the model from learning efficiently from input images and slows down the training process. To alleviate the effect of distance explosion while still take advantage of squared Euclidean distance being the regular Bregman divergence~\cite{banerjee2005clustering}, we propose to use leaky squared Euclidean distance to replace squared Euclidean distance as the metric to measure the similarity between features. 
In this way, the sparse gradients problem can be effectively solved by squashing the log-softmax activation into its non-linear area where the gradient is non-zero.
Our contributions can be summarized as follows.
\begin{itemize}
\item We provide insights of early sparse gradients problem when squared Euclidean distance and softmax activation are integrated in classification tasks.
\item We propose a simple yet effective scaling operation on Euclidean distance which we call it leaky squared Euclidean distance to mitigate the early sparse gradients problem.
\item We show that although early sparse gradients problem does not affect the convergence of the model alleviating the problem will improve classification accuracy in one-shot learning.
\end{itemize}

\section{Preliminary Knowledge}
\subsection{Softmax Activation}
\label{sec:softmax}
Softmax activation is also known as the normalized exponential function which transforms the $K$ dimensional input vector $\bm{f}$ to its exponential form and normalize each dimension of $\bm{f}$ by the summation of all exponential terms. Formally, softmax activation is defined as
\begin{equation}
\sigma(\bm{f}_i)=\frac{e^{\bm{f}_i}}{\Sigma^K_{j=1}e^{\bm{f}_j}}.
\end{equation}
$\bm{f}_i$ is the $i$-th dimension of the input vector $\bm{f}$. 
When softmax activation is applied to classifiers like the nearest neighbor classifier, $\bm{f}_i$ is replaced with the negative distance between the test image and its $i$-th neighbor. Then the probability of belonging to the nearest neighbor will be the largest among all the neighbors.

\subsection{One-shot Learning and Prototypical Networks}
\label{sec:oneshot}
One-shot learning is a task in which a classifier has to learn information about object categories based on only one labeled image per category. A straightforward way to yield a prediction is to compare the distances between the test image and the prototype image from each category. Then the test image is assigned to the category with smallest distance which is the same as the nearest neighbor rule. This is basically what prototypical networks~\cite{snell2017prototypical} do in few-shot learning. In prototypical networks, the prototype of each category is the mean feature vector within that category
\begin{equation}
\bm{\text{c}}_k=\frac{1}{|S_k|} \underset{(\bm{z}_i,y_i)\in S_k}{\sum} \bm{z}_i,
\end{equation}
where $S_k$ contains all feature vectors that belong to the $k$-th category. $\bm{z}_i$ and $y_i$ are the $i$-th feature vector and its corresponding label. $\bm{z}_i$ can be easily obtained from a convolutional neural network.
For each query feature vector $\bm{z}$, distances to all category prototypes are calculated as the input to softmax activation. 

\begin{equation}
p(y=y_i|\bm{z})=\frac{\text{exp}(-d(\bm{z}, \bm{c}_i))}{\sum^K_{j=1}\text{exp}(-d(\bm{z}, \bm{c}_j))}.
\label{eq:softmax_over_distance}
\end{equation}
$d(\bm{z}, c_i)$ indicates the squared Euclidean distance between $\bm{z}$ and $c_i$.
Then the network is trained via minimizing the following objective function
\begin{equation}
\ell = -\frac{1}{N}\sum\text{log}p(y=y_i|\bm{z}),
\label{eq:loss}
\end{equation}
where $N$ is the number of images in the mini-batch. The partial derivative with respect to $d(\bm{z}, \bm{c}_j)$ is 
\begin{equation}
\frac{\partial \ell}{\partial d(\bm{z}, \bm{c}_j)}=p(y_i=j|\bm{z})-1\{y_i=j\},
\label{eq:partial}
\end{equation}
where $1\{condition\}=1$ if $condition$ is satisfied and $1\{condition\}=0$ otherwise.

\begin{figure}
\centering
\includegraphics[width=0.8\linewidth]{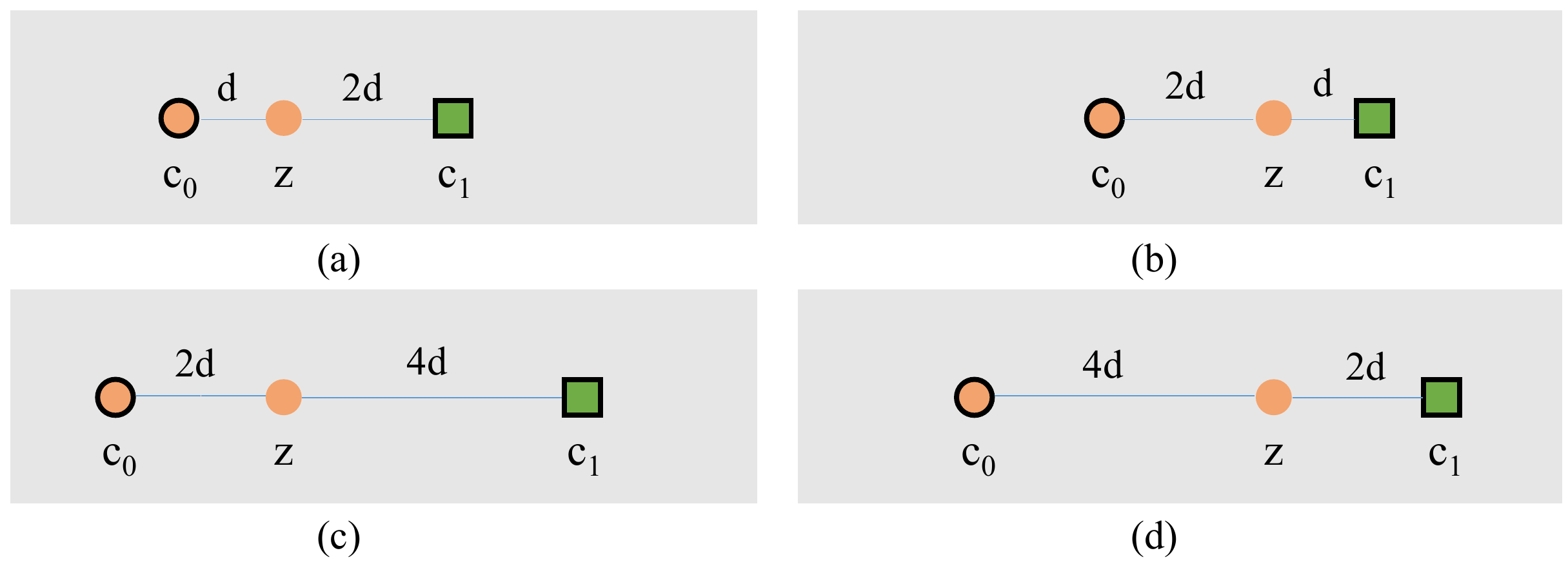}
%
\caption{
Scaling effect of softmax activation. If we simultaneously scale both $d(\bm{z}, \bm{c}_0)$ and $d(\bm{z}, \bm{c}_1)$ to a large magnitude, $p(y=0|\bm{z})$ will approach 1 in the left column while it will approach 0 in the right column. In both cases, $\frac{d(\bm{z}, \bm{c}_0)}{d(\bm{z}, \bm{c}_1)}$ remains unchanged.
(Best viewed in colors)
}
\label{fig:demo}
\end{figure}

\section{Early Sparse Gradients}
In order to illustrate the problem, we present a toy example in Figure~\ref{fig:demo}. Suppose we have an embedding function that can map three images to feature vectors $\bm{z}$, $\bm{c}_0$, and $\bm{c}_1$. $\bm{c}_0$ and $\bm{c}_1$ serve as the prototypes of two categories. $\bm{z}$ is the query image and its ground truth label is 0. The distances between $\bm{z}$ and $\bm{c}_0$, $\bm{c}_1$ are shown in Figure~\ref{fig:demo}. According to Equation~\ref{eq:softmax_over_distance}, 
it is obvious that $p_c(y=0|\bm{z})>p_a(y=0|\bm{z})$ and $p_d(y=0|\bm{z})<p_b(y=0|\bm{z})$. In more extreme cases, if $d(\bm{z}, \bm{c}_0)$ and $d(\bm{z}, \bm{c}_1)$ is multiplied by a large number, $p_c(y=0|\bm{z})$ will approach 1 and $p_d(y=0|\bm{z})$ will approach 0.  Then, their partial derivatives (see Equation~\ref{eq:partial}) will approach 0 and -1. It shows that even if $\frac{d(\bm{z}, \bm{c}_0)}{d(\bm{z}, \bm{c}_1)}$ remains unchanged, the sparsity of gradients can be totally different.
 
To give a more intuitive illustration, we plot curves of $-log(p(y=0|\bm{z}))$ with respect to three different $d(\bm{z}, \bm{c}_1)$ in Figure~\ref{fig:pa_wrt_b}. In area (I) and (II), we can find that the larger $d(\bm{z}, c_1)$ is, the earlier the saturation happens. When gradients are close to zero during back propagation, the intra-class variation will not decrease and the network learns little from training samples. 

On the contrary, if $d(\bm{z}, \bm{c}_1)$ is much smaller than $d(\bm{z}, \bm{c}_0)$, the curve enters into area (I), and the gradient will be close to -1. Though it is non-zero, if all incorrect predictions yield the same gradients, the model will ignore the diversity of the training data, causing the network to learn slowly.

Comparing the three curves, we also find that the non-linear area (area II) is becoming smaller when $d(\bm{z}, \bm{c}_1)$ is becoming larger. If $d(\bm{z}, \bm{c}_1)$ is large enough, the non-linear area will be negligible and thus the gradients will be either 0 or -1. 
One may argue that such large distance may not exist in real applications. However, suppose we are given two long vectors drawn from two random distributions. Calculating squared Euclidean distance will accumulate the squared difference of each dimension which increases the magnitude of the distance. In today's deep neural networks, the dimension of the output feature is usually very large. Distance explosion can be easily triggered and then early sparse gradients problem happens.

\section{Leaky Squared Euclidean Distance}
\label{leaky_squared Euclidean}
\begin{figure}
\centering
\subfigure[]{
  \label{fig:pa_wrt_b}
  \includegraphics[width=0.65\linewidth]{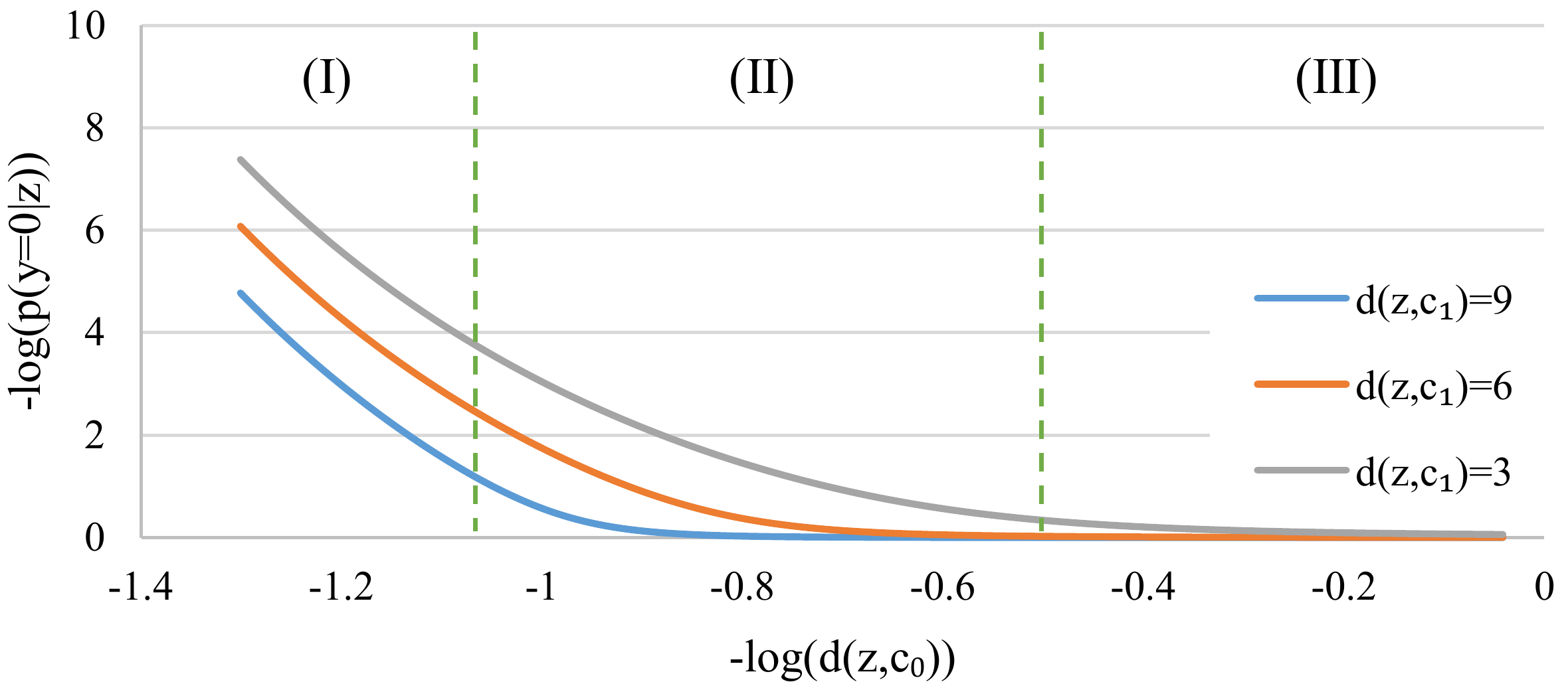}  
  }
\subfigure[]{
  \label{fig:leaky_dist}
  \includegraphics[width=0.32\linewidth]{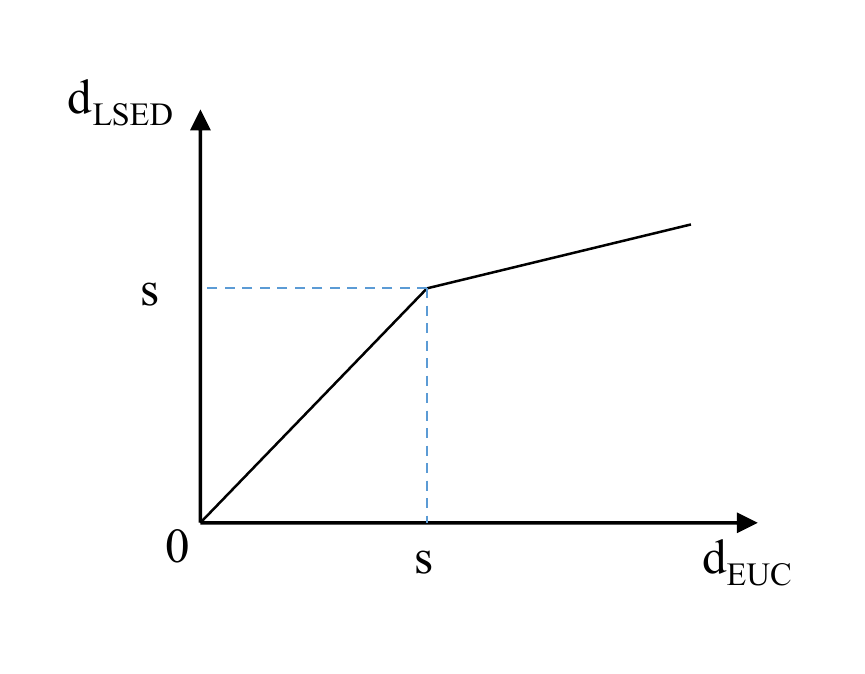}}
  
\caption{(a)$-log(p(y=0|\bm{z}))$ with respect to different $d(\bm{z}, c_1)$. In order to show different curves clearly, we choose $-log(d(\bm{z},c_0))$ instead of $-d(\bm{z},c_0)$ as the horizontal axis. (b) Leaky activation for squared Euclidean distance proposed in this work. (Best viewed in color.)}
\label{fig:pd_led}
\end{figure}

\begin{figure}
\centering
\subfigure[]{
  \label{fig:grad_hist_r1}
  \includegraphics[width=0.33\linewidth]{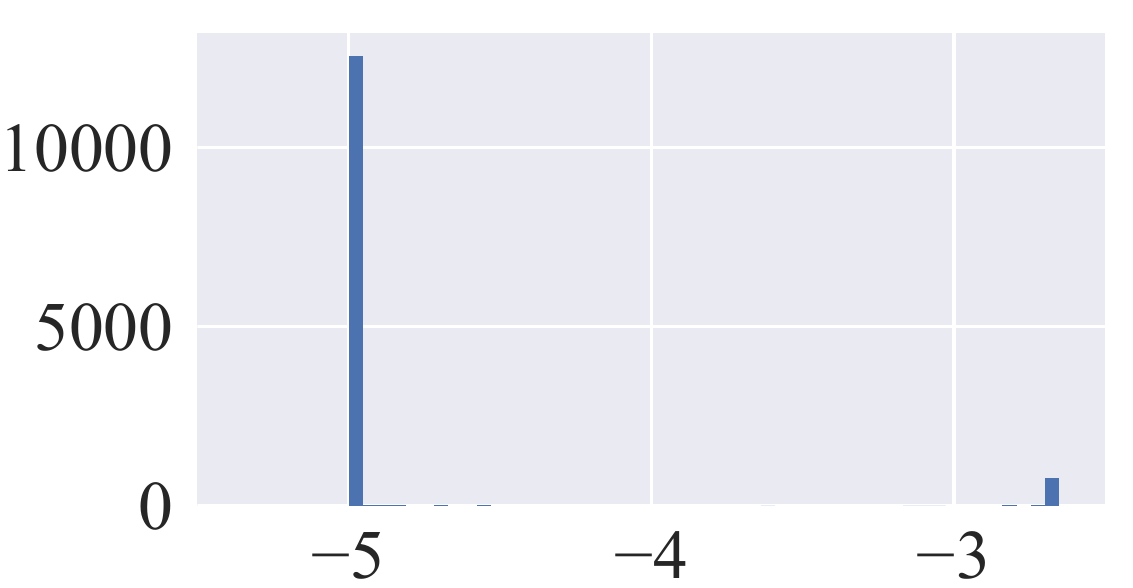}  
  }
\subfigure[]{
  \label{fig:grad_hist_r01}
  \includegraphics[width=0.33\linewidth]{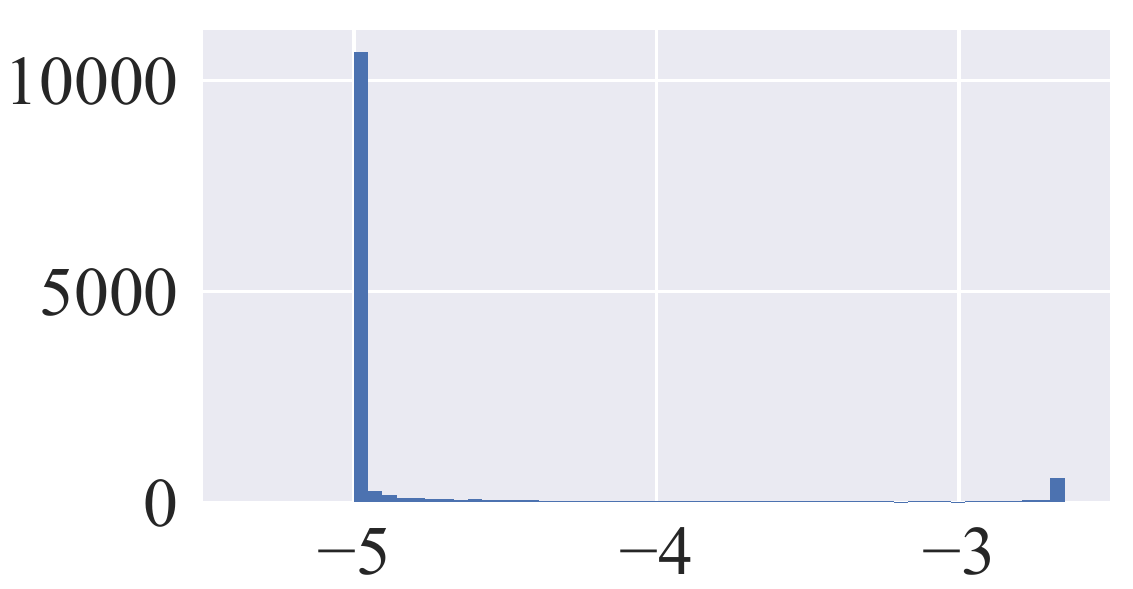}}
\subfigure[]{
  \label{fig:grad_hist_r001}
  \includegraphics[width=0.30\linewidth]{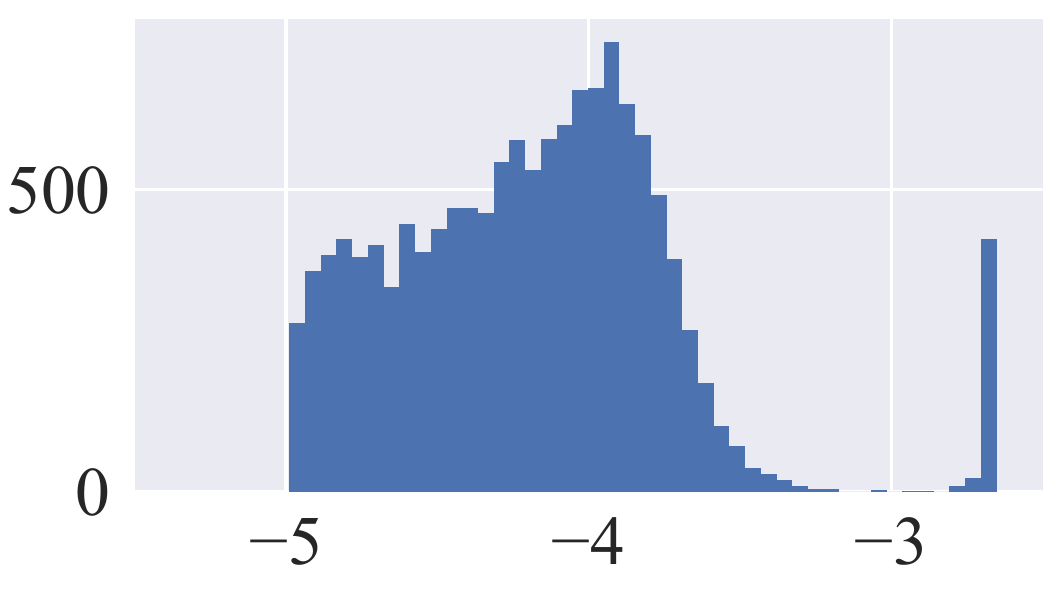}}
\caption{Histograms of $\text{log}(|\frac{\partial \ell}{\partial d(z, c_i)}|+10^{-5})$ from the network using (a) squared Euclidean distance, (b) leaky squared Euclidean distance with $s=0$ and $r=0.1$, and (c) leaky squared Euclidean distance with $s=0$ and $r=0.01$. 
}
\label{fig:grad_hist}
\end{figure}

Although sparse gradients do not harm the convergence of the model, if we take special care to alleviate the problem, we are likely to improve the final classification accuracy.  
To this end, we propose to use leaky squared Euclidean distance instead of squared Euclidean distance as the distance metric. The curve is shown in Figure~\ref{fig:leaky_dist}. It looks similar to Huber loss, but behaves differently. Formally, the leaky squared Euclidean distance is defined as 
\begin{eqnarray}
d_{LSED} = 
\begin{cases}
d_{EUC} & d_{EUC} <= s,\\
s + (d_{EUC}-s)r & d_{EUC}>s,
\end{cases}
\end{eqnarray}
where $s$ and $r$ are hyper-parameters. $d_{EUC}$ is the squared Euclidean distance. If $r=1$, leaky squared Euclidean distance will turn into squared Euclidean distance. So squared Euclidean distance is a special case of leaky squared Euclidean distance. However, if $r$ is chosen to be small enough, we can effectively reduce the distance to a small number so that distance explosion can be avoided and more gradients are squashed to be higher than zero but smaller than one. This behavior also connects the test sample and far-away category prototypes which do not contribute to the back propagation due to zero gradients when squared Euclidean distance is used.
We further show the comparison between histograms of $\text{log}(|\frac{\partial \ell}{\partial d(z, c_i)}|+10^{-5})$ using leaky squared Euclidean distance with different $r$ in Figure~\ref{fig:grad_hist}. Those gradients are calculated using prototypical networks for 1-shot 5-way classification task on miniImageNet dataset. Compared to Figure~\ref{fig:grad_hist_r1} in which gradients are sparse, leaky squared Euclidean distance with a small $r$ effectively alleviate the problem. When we decrease $r$ to a smaller value, the gradients will become less sparse.

%

\section{Experiments}
Since our work mainly concerns about the early sparse gradients problem when we apply softmax over squared Euclidean distances, we do not explore the whole space of network architectures. Improved classification accuracy compared to the baseline model would demonstrate the effectiveness of the proposed leaky squared Euclidean distance in a straightforward way. We also notice that there are other works that achieve promising results on few-shot learning~\cite{garcia2017few}\cite{mishra2017simple}. However, they either use much deeper network or complex training strategy. In this work, we choose recently proposed Prototypical Networks~\cite{snell2017prototypical} as our baseline model and compare the results using both squared Euclidean distance and leaky squared Euclidean distance.
The evaluation is performed on two benchmark datasets, i.e. Omniglot~\cite{lake2011one}\cite{lake2015human} and the miniImageNet version of ILSVRC-2012~\cite{russakovsky2015imagenet}. 

\subsection{Datasets}
\textbf{Omniglot dataset~\cite{lake2011one}\cite{lake2015human}} is a dataset containing 1623 different handwritten characters from 50 different alphabets. 
All images are resized to 28$\times$28. Each image is rotated three times in multiples of 90 degrees, yielding 6,492 classes in total. We follow the splits in \cite{ravi2016optimization} to split the dataset into 4,112 for training, 688 for validation, and 1,692 for testing. 

\textbf{MiniImageNet dataset~\cite{vinyals2016matching}} is a modified version of the ILSVRC-2012 dataset~\cite{russakovsky2015imagenet}. It contains 100 categories with 600 images per category. We follow the class split in~\cite{ravi2016optimization}. There are 64 classes for training, 16 for validation, and 20 for testing. All images are resize to 84$\times$84. No data augmentation is used during training and testing.

\subsection{Implementation Details}
Since the source code of prototypical networks is publicly available, we use the on-the-shelf implementation\footnote{https://github.com/jakesnell/prototypical-networks} from the authors. However, we could not reproduce the exact results presented in ~\cite{snell2017prototypical}. For a fair comparison, we will use the actual results from the authors' implementation run on Ominiglot dataset and miniImageNet dataset as our baseline results. 

For 1-shot and 5-shot training and testing, we follow the experiment protocols used in ~\cite{snell2017prototypical}. On Omniglot dataset, we perform 20-way classification for evaluation and on miniImageNet dataset we perform 5-way classification. 
The prototypical network contains four stacked blocks of \{3$\times$3-convolutional layer with 64 filters, batch-normalization, relu, 2$\times$2 max-pooling\}. For experiments on Omniglot dataset, we remove the last max-pooling layer so that the dimension of the resulting feature vector will be 576. In miniImageNet experiment, the dimension of the resulting feature vector is 1600.
The hyper parameters $s$ and $r$ are grid searched on the validation datasets. On Omniglot $s=0$ and $r=0.01$. On miniImageNet dataset $s=4$ and $r=0.01$. For the sake of simplicity, we will denote the prototypical networks using squared Euclidean distance as ProtoNet and prototypical networks using leaky squared Euclidean distance as ProtoNet(LSED). 

\begin{table}
\centering
\caption{20-way classification accuracy on Omniglot and 5-way classification accuracy on miniImageNet datasets using prototypical networks with squared Euclidean distance (ProtoNet) and leaky squared Euclidean distance (ProtoNet(LSED)).}
\label{tb:acc}
\begin{tabular}{lllll}
	\toprule
	              & \multicolumn{2}{l}{Omniglot} & \multicolumn{2}{l}{miniImageNet} \\
	\midrule
	Model         & 1-shot        & 5-shot       & 1-shot          & 5-shot         \\
	MatchingNet~\cite{vinyals2016matching}   & 93.5\%        & 98.7\%         & 46.6\%         & 60.0\%        \\
	Meta-Learner LSTM~\cite{ravi2016optimization} & $-$ & $-$ & 43.44 \% & 60.60\% \\
	ProtoNet~\cite{snell2017prototypical}      & 95.14\%       & 98.54\%         & 48.34\%         & 67.94\%        \\
	\midrule
	ProtoNet(LSED) & \bf{95.65}\%       & \bf{98.70}\%         & \bf{51.74}\%         & \bf{68.39}\% 		\\
	\bottomrule
\end{tabular}
\end{table}

\subsection{Omniglot One-shot Classification}
\begin{figure}
\centering
\includegraphics[width=0.98\linewidth]{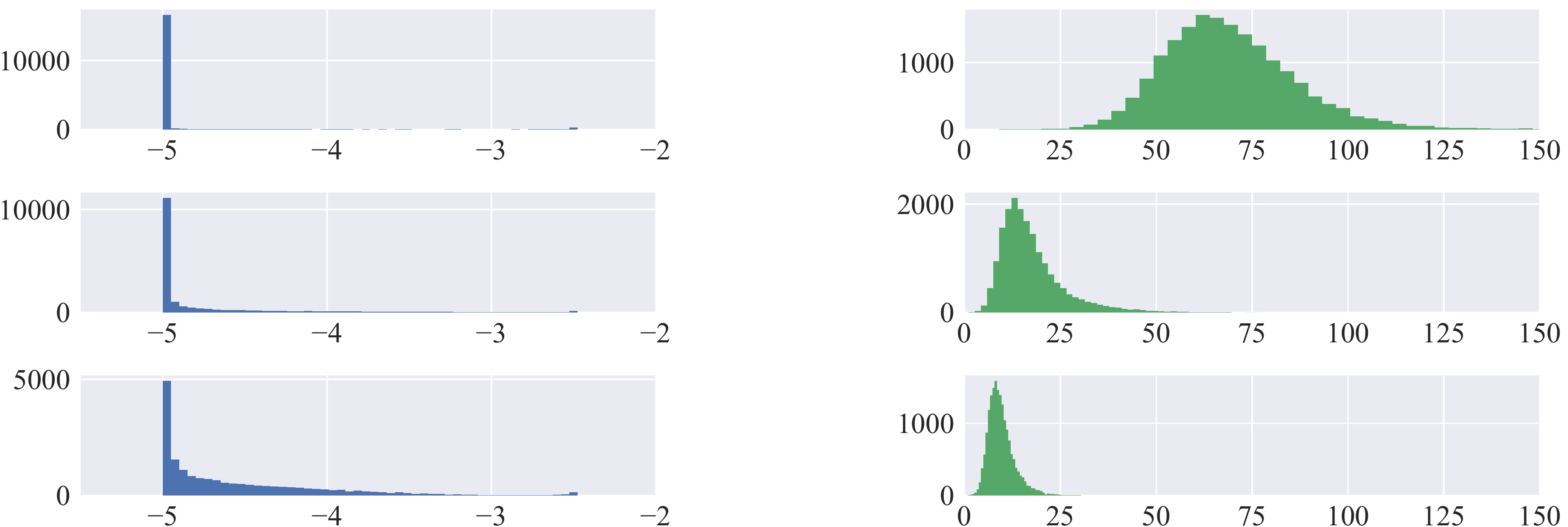}
\caption{Histograms of $\text{log}(|\frac{\partial \ell}{\partial d(z, c_k)}|+10^{-5})$ (left column) and distances (right column) between query images and category prototypes after 0 (1st row), 8 (2nd row) and 16 (3rd row) training iterations on Omniglot dataset. The bias term $10^{-5}$ is added for the purpose of better illustration. The model used here is ProtoNet.}
\label{fig:glot_no_last_pool_base}
\end{figure}

\begin{figure}
\centering
\includegraphics[width=0.98\linewidth]{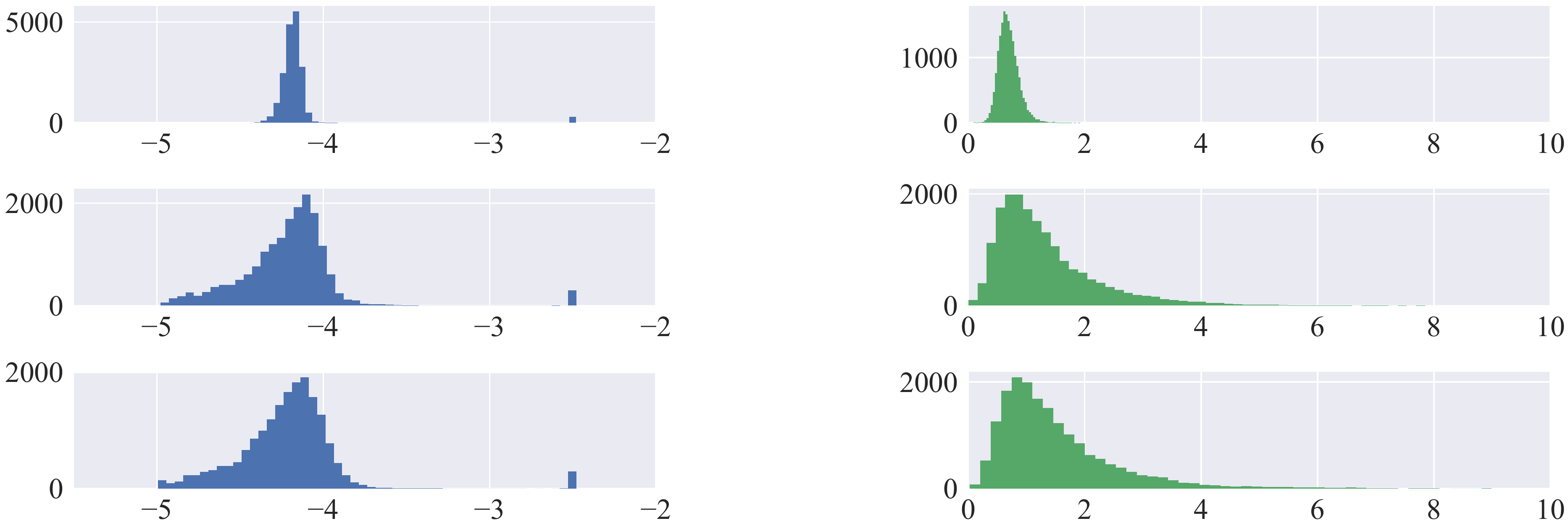}
\caption{Histograms of $\text{log}(|\frac{\partial \ell}{\partial d(z, c_k)}|+10^{-5})$ (left column) and distances (right column) between query images and category prototypes after 0 (1st row), 8 (2nd row) and 16 (3rd row) training iterations on Omniglot dataset. The bias term $10^{-5}$ is added for the purpose of better illustration. The model used here is ProtoNet(LSED).}
\label{fig:glot_no_last_pool_leaky}
\end{figure}

In this section, we show our results on the task of 1-shot 20-way classification on Omniglot dataset. The classification accuracy of ProtoNet(LSED) is 95.65\% versus 95.14\% using ProtoNet as shown in Table~\ref{tb:acc}.
The histograms of $\text{log}(|\frac{\partial \ell}{\partial d(z, c_k)}|+10^{-5})$ of ProtoNet and ProtoNet(LSED) are shown in Figure~\ref{fig:glot_no_last_pool_base} and Figure~\ref{fig:glot_no_last_pool_leaky}. We also show the histograms of distances between query images and category prototypes by the side of the histograms of log-gradients.
Since $|\frac{\partial \ell}{\partial d(z, c_k)}|$ can be very close to zero, we add a bias term $10^{-5}$ to make the gradients slightly larger than zero. 

In Figure~\ref{fig:glot_no_last_pool_base}, most of the gradients are close to zero. As training proceeds, a few more gradients become nonzero after 8 iterations and more after 16 training iterations. This phenomena can be explained by the fact that the network is initialized randomly therefore squared Euclidean distances between query images and other category prototypes are very large (see the right column in Figure~\ref{fig:glot_no_last_pool_base}). Then the early sparse gradients problem is triggered.
However, when the network is trained for a few iterations, the embedding becomes more semantic, and thus the distances between images become smaller, pushing the negative log-softmax into the non-linear area(area (II) in Figure~\ref{fig:pa_wrt_b}). 

When we change squared Euclidean distance to leaky squared Euclidean distance, we can see from Figure~\ref{fig:glot_no_last_pool_leaky} that distances between images are much smaller than those in Figure~\ref{fig:glot_no_last_pool_base} in the first training iteration. Therefore, most of the gradients are non-zero. After training for 16 iterations, both the distances and the gradients become more diverse which helps the network to learn better knowledge of categories.

\begin{figure}
\centering
\includegraphics[width=0.98\linewidth]{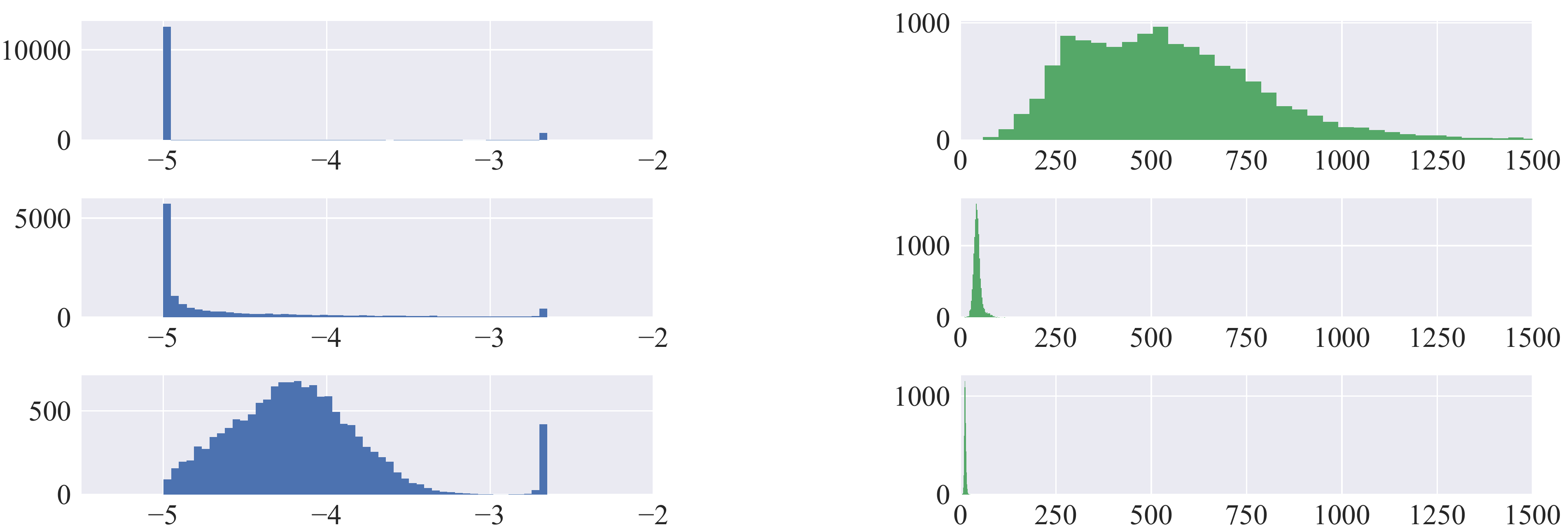}
\caption{Histograms of $\text{log}(|\frac{\partial \ell}{\partial d(z, c_k)}|+10^{-5})$ (left column) and distances (right column) between query images and category prototypes after 0 (1st row), 50 (2nd row) and 100 (3rd row) training iterations on miniImageNet dataset. The bias term $10^{-5}$ is added for the purpose of better illustration. The model used here is ProtoNet.}
\label{fig:mini_base}
\end{figure}

\begin{figure}
\centering
\includegraphics[width=0.98\linewidth]{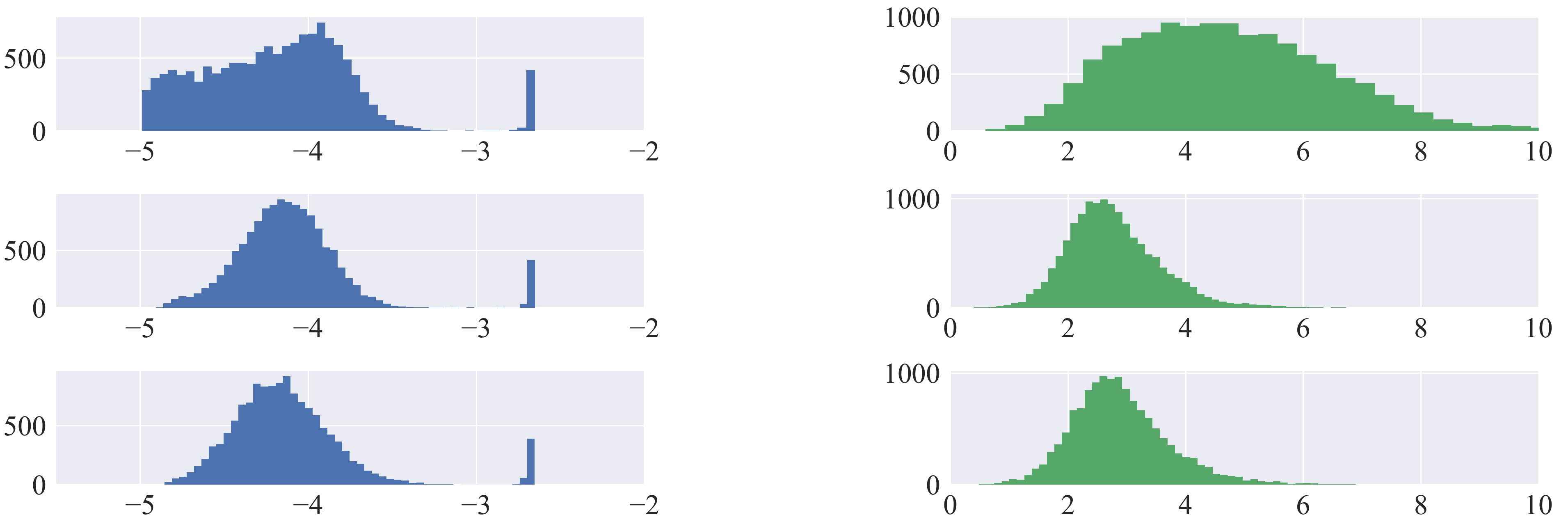}
\caption{Histograms of $\text{log}(|\frac{\partial \ell}{\partial d(z, c_k)}|+10^{-5})$ (left column) and distances (right column) between query images and category prototypes after 0 (1st row), 50 (2nd row) and 100 (3rd row) training iterations on miniImageNet dataset. The bias term $10^{-5}$ is added for the purpose of better illustration. The model used here is ProtoNet(LSED).}
\label{fig:mini_leaky}
\end{figure}

\begin{figure}
\centering
\subfigure[]{
  \label{fig:var_s}
  \includegraphics[width=0.48\linewidth]{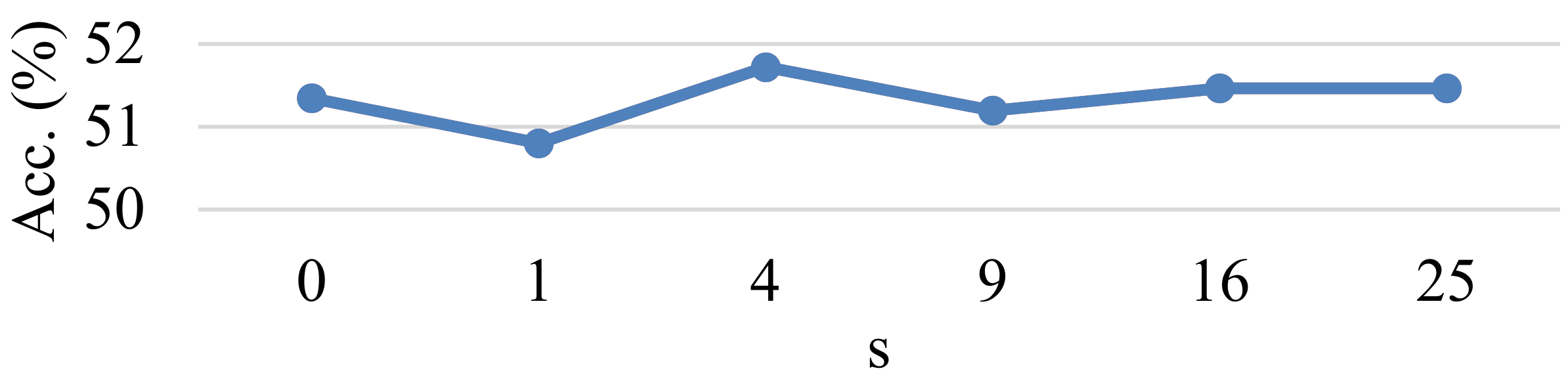}}
\subfigure[]{
  \label{fig:var_r}
  \includegraphics[width=0.48\linewidth]{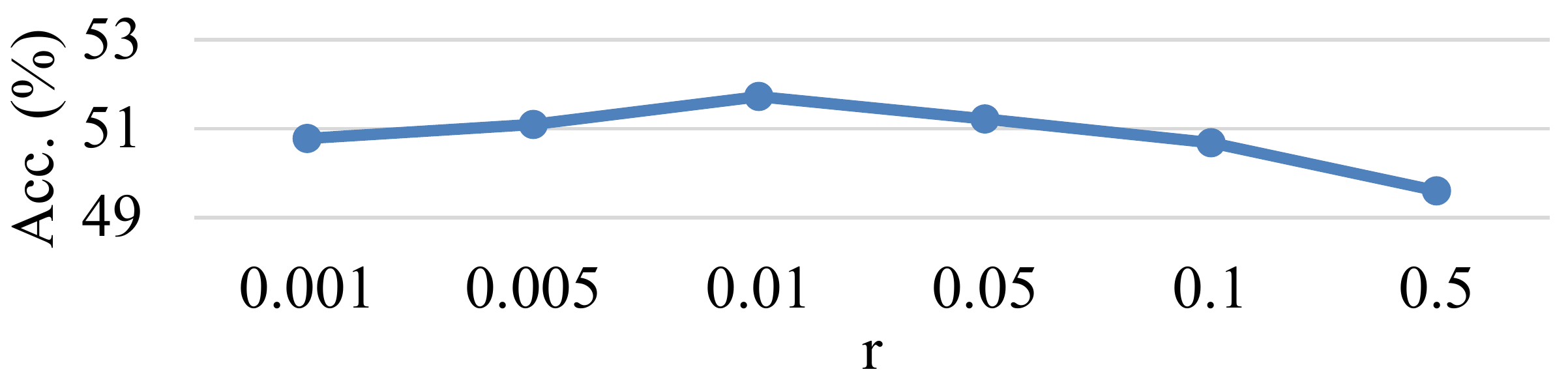}}
\caption{Evaludation of the effect of $s$ and $r$ on the task of 1-shot 5-way classification on miniImageNet dataset. In (a), we fix $r=0.01$ and train ProtoNet(LSED) with different $s$. In (b), we fix $s=4$ and train ProtoNet(LSED) with different $r$. }
\label{fig:var_s_r}
\end{figure}

\subsection{MiniImageNet One-shot Classification}
In this section, we show the results on miniImageNet dataset. The classification accuracy is 51.74\% from ProtoNet(LSED) versus 48.34\% from ProtoNet as shown in Table~\ref{tb:acc}. The evolve of log-gradient histograms and distance histograms are shown in Figure~\ref{fig:mini_base} and Figure~\ref{fig:mini_leaky}. 

From Figure~\ref{fig:mini_base}, we can find that at the initial point the gradients are sparse due to the extreme large distances between query images and category prototypes. 
Note that squared Euclidean distances are as large as 1,500 which is much larger than those in Omniglot experiment. This is attribute to two aspects. Firstly, the dimension of the feature vector is 1600 which is much larger than 576 used in Omniglot experiment. Secondly, images in miniImageNet dataset are also more complex than those in Omniglot. In the initial training stage, the number of non-zero gradients is small which prevents the network from efficient knowledge learning. 
When the network is trained for 100 iterations, the gradients are less sparse. The distance histograms also show a sharp drop in magnitude. It indicates that squared Euclidean distance is far from an optimal similarity metric in the initial training stage. 

In Figure~\ref{fig:mini_leaky}, the initial gradients are not sparse any more so that the network can learn quickly from different training images. The distance histograms do not change drastically either, indicating the superiority over squared Euclidean distance.

\begin{figure}
\centering
\includegraphics[width=0.88\linewidth]{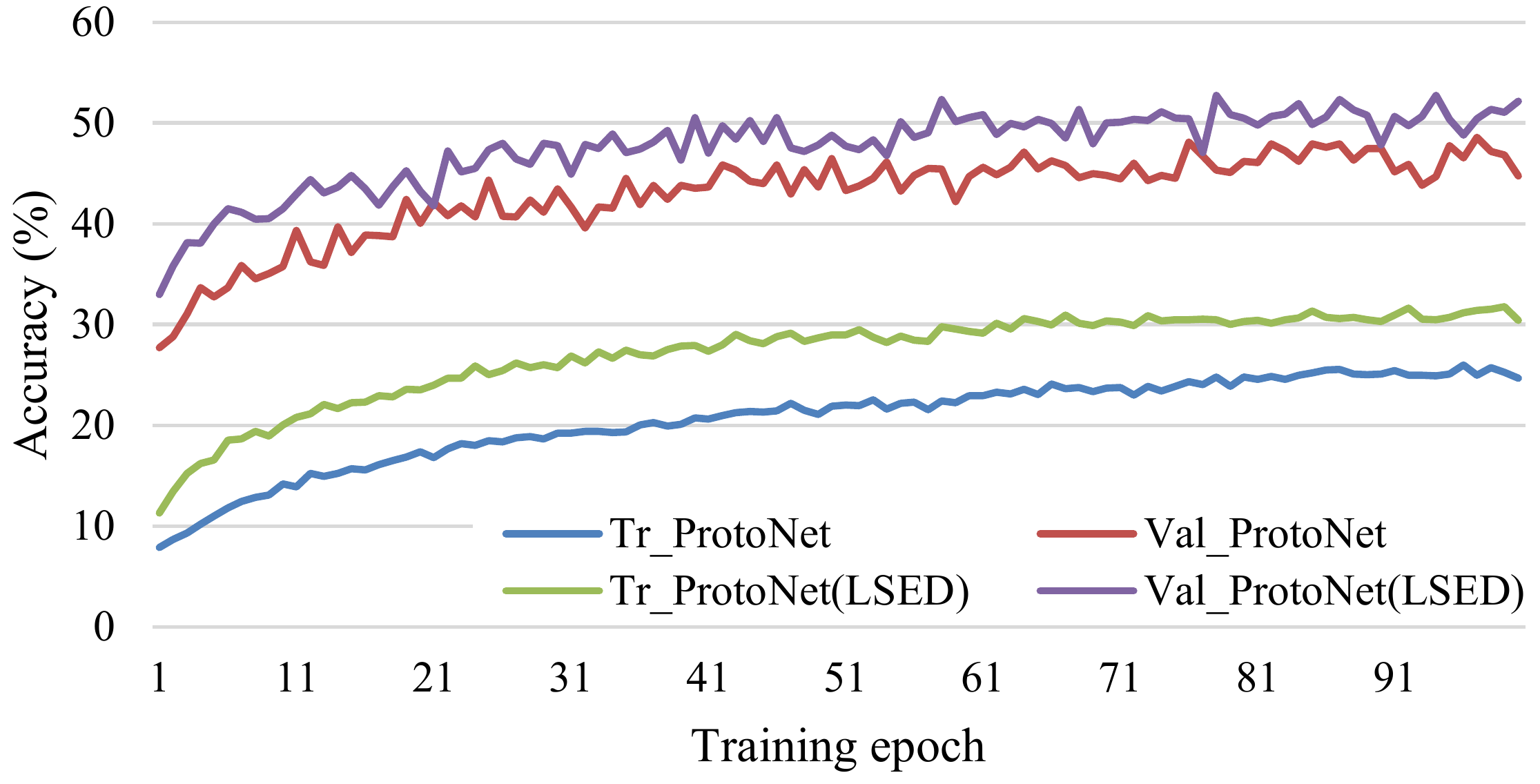}
\caption{Training/validation curves of PtotoNet and ProtoNet(LSED) on the task of 1-shot 5-way classification on miniImageNet dataset. \lq\lq Tr\_ProtoNet\rq\rq ~and \lq\lq Tr\_ProtoNet(LSED)\rq\rq ~mean the training accuracy from the two models while \lq\lq Val\_ProtoNet\rq\rq ~and \lq\lq Val\_ProtoNet(LSED)\rq\rq ~indicate the validation accuracy. (Best viewed in color.)}
\label{fig:acc_cmp}
\end{figure}

In Figure~\ref{fig:acc_cmp}, we also show the training and validation curves of ProtoNet and ProtoNet(LSED) on the task of 1-shot 5-way classification on miniImagenet dataset. The training accuracy is lower than the validation accuracy. This is because we follow the training protocols in ~\cite{snell2017prototypical} in which the training task is designed to be more difficult than the validation task. From the figure, we can clearly observe that after the first training epoch both training accuracy and validation accuracy of ProtoNet(LSED) are higher than those of ProtoNet. It provides evidence that leaky squared Euclidean distance helps improve learning speed. The large margin of classification accuracy consistently existing between the two models also demonstrates the effectiveness of the proposed leaky squared Euclidean distance.

To evaluate the effect of $s$ and $r$ in leaky squared Euclidean distance, we perform two more experiments. The results are shown in Figure~\ref{fig:var_s_r}. In Figure~\ref{fig:var_s}, we fix $r=0.01$ and plot the test accuracy from model trained with different $s$. Results on the test set are within $\sim$1\% range indicating the stable performance of our leaky squared Euclidean distance with respect to $s$. In Figure~\ref{fig:var_r}, we fix $s=4$ and observe that if $r$ is very close to 1, the test accuracy drops. It is as expected because leaky squared Euclidean distance turns into squared Euclidean distance when $r=1$.

\subsection{Few-shot Learning}
We also conduct 5-shot classification experiments on both Omniglot dataset and miniImage dataset. The classification accuracy is shown in Table~\ref{tb:acc}. We observe only marginal accuracy improvement. The reason is that in the scenario of 5-shot learning, category prototypes are the averaged feature representation of labeled samples which removes much of the intra-class variations. Thus the distance explosion is less likely to happen which is helpful for alleviating the early sparse gradients problem. 

\section{Related Work}
There are several variants of softmax activation.
In~\cite{wen2016discriminative}, Yandong et al. show that softmax loss does not preserve the compactness of clusters in image classification. They propose a center loss which explicitly pulls features within a cluster to the cluster center. 
In~\cite{liu2016large}, Weiyang et al. find that there is no constraint for the margins between clusters in softmax loss. They present a modification of softmax activation to maximize the margins.
In~\cite{liu2017sphereface}, Weiyang et al. introduce the angular softmax (A-Softmax) loss which can be viewed as imposing discriminative constraints on a hypersphere manifold to enable convolutional neural networks to learn angularly discriminative features.
In~\cite{chen2017noisy}, Binghui et al. show that the individual saturation leads to short-lived gradients propagation in softmax activation which is poor for robust exploration of SGD. They suggest to use annealed noise injection to mitigate this problem.


\section{Conclusion}
In this paper, we investigate the problem of early sparse gradients when applying softmax activation over distances. We proposed to use leaky squared Euclidean distance to replace squared Euclidean distance to rescale the distance to a smaller magnitude so that gradients can be pushed to a more diverse area. Though we mainly focus on one-shot learning in this work, we provide insights of designing distance metric for softmax activation which can be helpful for other related tasks.

\bibliographystyle{ieee}
\bibliography{ref}

\end{document}